\documentclass[runningheads]{llncs}
\usepackage{graphicx}

\usepackage{tikz}
\usepackage{comment}
\usepackage{amsmath,amssymb} 
\usepackage{color}
\usepackage{tabularx}
\usepackage{multirow}
\usepackage{caption}
\usepackage{booktabs}
\usepackage{hyperref}
\usepackage{cite}
\usepackage[accsupp]{axessibility}  
\usepackage{caption}
\usepackage{subcaption}

\usepackage{xcolor}
\usepackage{gensymb}

\definecolor{mygray}{RGB}{128, 128, 128}

\usepackage{enumitem}

\newcommand{\etal}{\textit{et al.}}
\newcommand{\ie}{\textit{i.e. }}
\newcommand{\eg}{\textit{e.g. }}

\newcolumntype{k}{>{\hsize=5\hsize}X}
\newcolumntype{w}{>{\hsize=4\hsize}X}
\newcolumntype{l}{>{\hsize=4\hsize}X}

\definecolor{Stu}{RGB}{0,148,255}
\definecolor{Ale}{RGB}{200,48,0}
\definecolor{Taiana}{RGB}{145,6,106}
\definecolor{Toso}{RGB}{148,190,50}

\begin{document}

\pagestyle{headings}
\mainmatter
\def\ECCVSubNumber{7218}  

\title{PoserNet: Refining Relative Camera Poses Exploiting Object Detections \thanks{This project has received funding from the European Union's Horizon 2020 research and innovation programme under grant agreement No 870743.}}

\titlerunning{PoserNet}
%
\author{Matteo Taiana \and                                    
Matteo Toso \and                                         
Stuart James\and                                         
Alessio Del Bue}
\authorrunning{M. Taiana~\etal}
%
\institute{Pattern Analysis and Computer Vision (PAVIS), Istituto Italiano di Tecnologia (IIT),
Genoa, Italy \\
\email{\{name.surname\}@iit.it}}
\maketitle
\begin{abstract}
The estimation of the camera poses associated with a set of images commonly relies on feature matches between the images. In contrast, we are the first to address this challenge by using objectness regions to guide the pose estimation problem rather than explicit semantic object detections. We propose Pose Refiner Network (PoserNet) a light-weight Graph Neural Network to refine the approximate pair-wise relative camera poses. PoserNet exploits associations between the objectness regions - concisely expressed as bounding boxes - across multiple views to globally refine sparsely connected view graphs.
We evaluate on the 7-Scenes dataset across varied sizes of graphs and show how this process can be beneficial to optimisation-based Motion Averaging algorithms improving the median error on the rotation by 62\degree with respect to the initial estimates obtained based on bounding boxes. Code and data are available at \href{https://github.com/IIT-PAVIS/PoserNet}{github.com/IIT-PAVIS/PoserNet.}
\end{abstract}

\section{Introduction}
A common problem in computer vision is the recovery of  multiple camera poses in
a common reference frame starting from a set of images, with applications in tasks such 
as Structure from Motion (SfM), Simultaneous Localisation And Mapping (SLAM) or visual odometry.
Traditionally, most solutions rely on the extraction and matching of keypoint features from the images~\cite{schoenberger2016sfm}.
Those approaches are vulnerable to many factors, such as: changes in viewpoint, illumination, repeated patterns that can lead to mismatches; and the presence of featureless, transparent or reflective surfaces which can result in a scarcity of useful keypoints.
Since all these factors commonly occur in real-world scenes, feature-based approaches often rely on 
extensive refinement steps of the matches, and on outlier rejection methods like RANSAC~\cite{fischler1981random}.

\begin{figure*}[t]
    \centering
    \includegraphics[width=\linewidth]{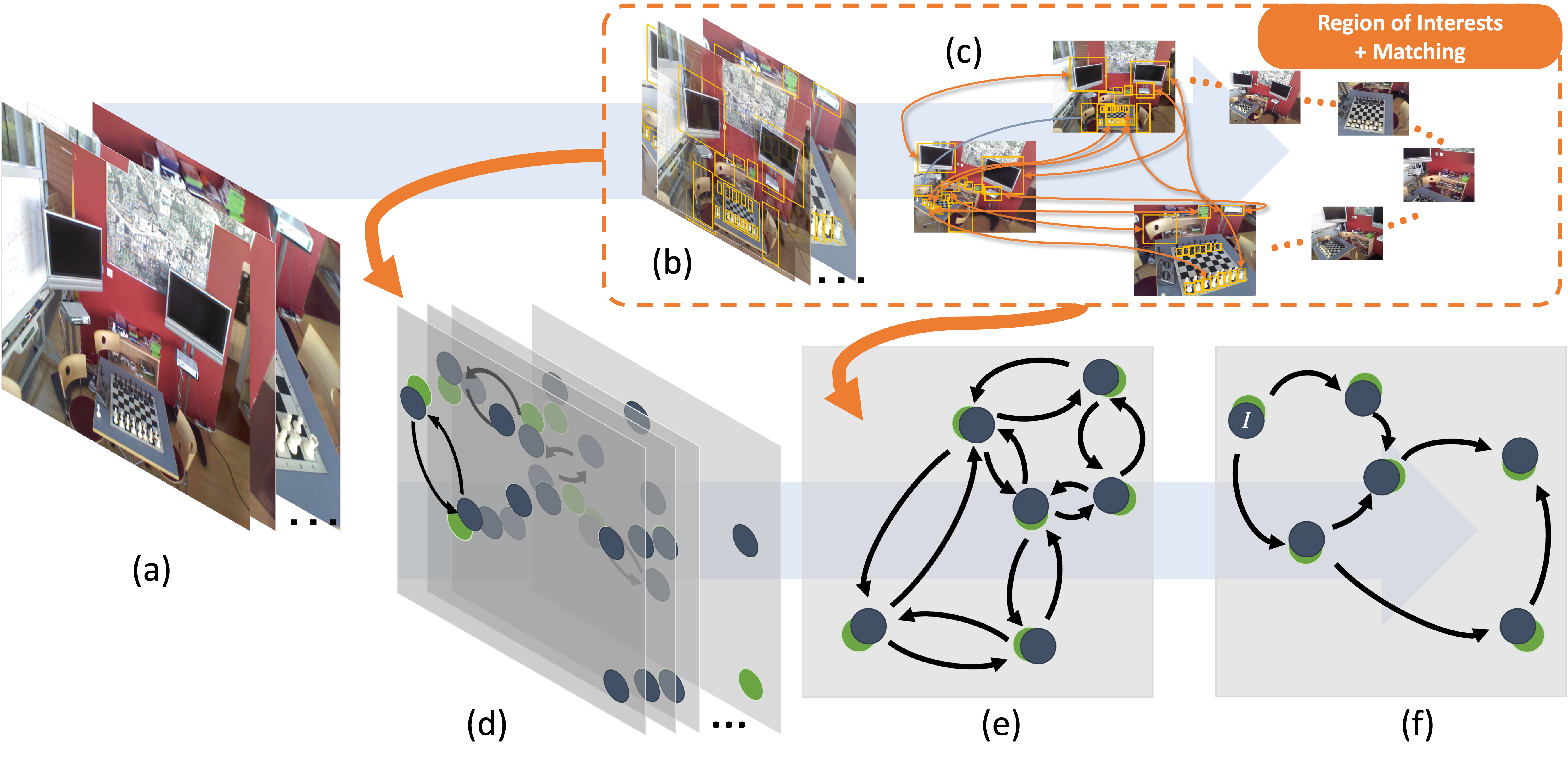}
    \caption{(a) Our method takes a set of unordered images as input; (b) it detects ROIs; and (c) matches them between frames; (d) ROI \& matches are then used to estimate an initial relative pose using 5-point algorithm; (e) the initial poses are refined exploiting the matched detections by PoserNet; and, finally, (f) the refined poses are processed by Motion Averaging to compute absolute poses.}
    \label{fig:pipeline}
\end{figure*}

We propose an alternative approach, instead of focusing on keypoints, we leverage on the
continuous improvements in object detectors~\cite{renNIPS15fasterrcnn,kim2021oln} to extract the location of potential objects in the images as the basis of our approach. This presents multiple advantages: \emph{i)} detections matching is more reliable; \emph{ii)} working
with objects is more efficient than handling tens of thousands of keypoints; and
\emph{iii)} for many downstream applications it is convenient to reason in terms of objects present in the
scene, rather than 3D points. The use of objects for camera pose estimation, \ie. Structure from Motion with Objects (SfMO), has been previously proposed~\cite{crocco2016structure,gay2017probabilistic}, however, it relies on a closed-form solution based on multiple disparate views of objects and a simpler camera model that makes its application highly constrained and challenging to apply in the wild. 

To overcome the limitation of SfMO and test our object-based assumptions,  we propose: \emph{i)} to extract a set of 2D bounding boxes using an objectness detector indicating the probable location of objects of interest; \emph{ii)} to match these detections across images; \emph{iii)} to use them to provide an initial estimate of the relative transformation between each pair of images that contain common scene elements;
 and \emph{iv)} to exploit the matched detections and rough relative poses to create a view graph, which we then refine via 
a novel GNN-based architecture called Pose Refinement Network (PoserNet). This allows us to obtain more accurate relative camera poses,
which \emph{v)} we then use to recover absolute poses via Motion Averaging algorithms and place camera poses in a common reference frame (as seen in Fig.~\ref{fig:pipeline}).

\noindent To summarise, the contributions of this paper are threefold:
\begin{itemize}
\item A GNN-based architecture, PoserNet, that exploits matched region detections to refine rough estimates of the pair-wise camera
transformations, producing accurate relative poses; 
\item we combine the PoserNet module with two paradigms of Motion Averaging algorithms (EIG-SE3\cite{arrigoni2016spectral} \& MultiReg\cite{yew2020-RobustSync}) to estimate absolute camera poses;
\item we provide experimental evidence using object detections and keypoints, 
comparing our method against a state-of-the-art methods for small 
view-graphs on 7-Scenes.
\end{itemize}

\section{Related work}
Recovering camera poses from a sparse set of images is a well studied problem,
especially in the context of SfM, and there are extensive reviews on the topic (see Bianco~\etal~\cite{bianco}).
Similarly, many works discussed in this paper, and our PoserNet model, are based on
Graph Neural Networks (GNN). These have recently received a lot of attention, due to their successful applications in many domains, and we point to Wu~\etal~\cite{Wu2021ACS} and Zhou~\etal~\cite{zhou2020graph} for an extensive review of GNNs.
In this section, we will instead discuss: methods for relative camera pose estimation (Sec.~\ref{sec:relativeposelit}), as these are the approaches PoserNet aims to improve on; Motion Averaging works (Sec.~\ref{sec:motionavglit}), which allow us to evaluate the effects of PoserNet pose refinements; and object-based approaches in 3D Computer Vision (Sec.~\ref{sec:objectincvlit}), as PoserNet takes as inputs matched object detections. 

\subsection{Relative Pose Estimation}
\label{sec:relativeposelit}
The most common approach to relative pose estimation is to 
extract a large number of keypoints, matched across the images, and to apply geometrical methods like
Hartley's normalised eight-point algorithm~\cite{601246}, Nist\'er's five-point algorithm~\cite{1288525} 
or its simplified implementation introduced by Li and Hartley~\cite{ICPR2006579}. 
This problem is hard in real scenarios, since pixel locations and camera extrinsics can assume a wide range of values, and keypoint matches are noisy and unreliable in many practical application. For this reason, such geometrical methods are typically paired with tools like RANSAC to reduce the impact of outliers.
The advancement of deep learning allowed approaching relative pose estimation as a learning problem:  
Li~\etal~\cite{LI2021134} directly trained a deep network to predict the relative pose and orientation between two images, in the context of image relocalisation.
Moran~\etal~\cite{moran2021deep} directly recovered global camera poses from feature matches and a sparse scene reconstruction, minimising an unsupervised reprojection loss exploiting one encoder equivariant to feature permutations. 
Cai~\etal\cite{Cai2021Extreme} proposed an encoders-based network, with dense correlation volume representation,
trained to estimate the relative pose from image pairs via cues like light source directions, vanishing points, and scene symmetries.
Alternatively, Neural Radiance Fields for View Synthesis (NeRF)~\cite{mildenhall2020nerf}
was re-framed by Yen-Chen~\etal~\cite{yen2020inerf} and used for camera pose estimation. 
Yan~\etal~\cite{bs200908049}, used a GNN to frame the problem of image retrieval for camera relocalisation as a node
binary classification in a subgraph. 

Unlike these methods, we suggest that the relative poses estimated via geometrical approaches - which are typically computationally efficient and do not require training - can provide competitive performance, if properly refined. We exemplify this by showing how even rough estimates obtained by applying a standard geometrical method to a small number of loosely matched points (the centers of matched detections) can be refined by our PoserNet.

\subsection{Motion Averaging}
\label{sec:motionavglit}
Given a set of relative poses, there are different approaches to reconstruct absolute poses in a common reference frame
(\ie the Motion Averaging or synchronisation problem).
Classically, this is done via optimisation algorithms, 
like the ones of standard global SfM~\cite{moulon2016openmvg}, or of Arrigoni~\etal~\cite{arrigoni2016spectral}, who proposed a closed-form solution based on spectral decomposition, using an iteratively reweighted Least Squares scheme to reduce the impact of outliers. Recently, Arrigoni~\etal~\cite{Arrigoni_2021_ICCV} showed how cycle consistency allows establishing whether a view graph is solvable or not.
Lee and Civera~\cite{Lee_2021_CVPR} proposed a rotation-only Bundle Adjustment (BA) framework, decoupling rotation estimation from translation and scene reconstruction, and
Chen~\etal~\cite{9577752} combined a global optimiser with fast view graph filtering and a local optimiser, improving on the traditional BA loss.

Also in the case of Motion Averaging, there has been a proliferation of deep models, such as
NeuRoRA~\cite{purkait2020neurora}, which employed two GNNs to clean the relative camera poses before computing absolute poses.
Yang~\etal~\cite{Yang_2021_CVPR} introduced an end-to-end neural network for multiple rotation averaging and optimisation. Its view-graph is refined by fusing image context and features to predict outlier rotations, and initialising the rotation averaging process through a differentiable module, that allows outlier refinement at test time.
Similarly, Yew and Lee~\cite{yew2020-RobustSync} used a GNN to learn transformation synchronisation, but use an iterative approach in the tangent space where each step consists of a single weight-shared message passing layer that refines the absolute poses. 

We test the effects of relative pose refinement via PoserNet on one optimisation-based~\cite{arrigoni2016spectral} and one deep model~\cite{yew2020-RobustSync}, as both cases present interesting advantages. Classical methods do not require training, and therefore generalise better to different use cases, and are typically efficient, easy to deploy and well-studied; on the other hand, successfully combining PoserNet with a deep learning method would allow merging them in an end-to-end pipeline.
While PoserNet shares some similarities with methods like~\cite{purkait2020neurora}, we focus on refining the relative poses before the Motion Averaging task, providing better inputs. Moreover, our method reasons about all relative poses of the graph while also exploiting bounding box information usually not used by models like~\cite{purkait2020neurora}(\eg the location of the object detections).

\subsection{Object-Based Computer Vision}
\label{sec:objectincvlit}
One of our key insight is the use of object locations instead of generic keypoints, as initial step of the camera pose reconstruciton process.
Other works have already suggested 
reasoning about the world in terms of object detections: 
Rubino~\etal~\cite{rubino2017pami} proposed a closed-form solution for estimating the 3D occupancy of an object,
from multiview 2D object detections, and Gaudilliere~\etal~\cite{camrelocellipse} used a similar representation to show how,
in a scene represented as a collection of ellipsoids, a single ellipse-ellipsoid match and an approximate orientation from IMUs or vanishing point algorithms is enough to recover the camera location. They later~\cite{8968180} showed how, given two ellipse-ellipsoid correspondences, the camera rotation is recovered via an optimisation problem over the three Euler angles.  The problem can then be further simplified~\cite{9127873}, reducing it to an optimisation
over a single angle by introducing constraints satisfied in many
real world applications, \ie that the images
have null azimuth angle, and that the segment connecting the centres of the ellipsoids projects on the one connecting the centres of the detected ellipses. 

In the context of object-level SLAM, notable contributions include Fusion++ ~\cite{McCormac2018FusionVO}, which combines Microsoft Fusion's~\cite{6162880} TSDF and instance segmentation for indoor scene understanding; or SLAM++~\cite{6619022}, which implements real-time object detection, tracking and mapping in a dense RGB-D framework, but requires a database of well defined 3D models of the objects; or QuadricSLAM~\cite{quadslam}, which uses noisy odometry and object detections (as bounding boxes) to simultaneously estimate camera poses and objects locations, expressed as ellipsoids.

Unlike the aforementioned methods, we do not use a specific parameterisation of the
objects shape (\eg ellipsoid), and only rely on the approximate location of generic ROIs. Moreover, our method can be seen as complementary to theirs, as in principle we could take the bounding boxes associated with our ROIs, and the camera poses produced by combining PoserNet with a Motion Averaging approach, and use one of 
the approaches to localise the scene elements in 3D. Regarding SLAM methods, a direct comparison is not possible, as our approach is aimed at scenes with large camera baseline displacement and sparsely connected view-graphs, with minimal overlapping views. 

\section{Methodology}
Our approach takes as input an unordered set of images $\{I_i\}$, $i \in [1,...,N]$ and their respective camera matrix intrinsics (${K_i}$). We then aim to reconstruct the extrinsics $M_i=[R_i|t_i]$, with rotation $R_i\in SO(3)$ \ie Special Orthogonal Group, and translation $t_i\in \mathbb{R}^{3}$ that map from the world reference frame to the camera reference frame (as shown in Figure~\ref{fig:pipeline}). Firstly we detect ROIs within $I_i$, then we perform matching between ROIs and in turn solve for an initial estimate of the pairwise problem (Sec.~\ref{sec:det_mat}). The ROIs represented as bounding boxes (BB) and the initial estimates are then passed to our PoserNet module to refine the relative poses (Sec.~\ref{sec:poser}). Finally, the refined relative poses are processed by a rotation averaging method, which computes the absolute poses estimates of the cameras (Sec.~\ref{motionaveragingexplained}).

\subsection{Objectness, matching and initial poses }\label{sec:det_mat}
For each image, we compute a set of bounding boxes $bb_{i,a}$, $a\in [1,...,50]$ which represent the top most confident ROIs candidates for objects. We opt for ROIs instead of the explicit object detections, as we make no assumption on semantic classes and they provide bounding boxes even in scenes with few objects. We use the \textit{objectness}-based detector, Object Localisation Network (OLN)~\cite{kim2021oln}. OLN is a generalisation of common detectors, substituting the classification layer (object semantic class) with a localization confidence and trained in a weakly supervised manner relying on intersection over union and centerness. 

Once we have a set of bounding boxes for each image, we then match them to obtain tracks of detections across the multi-view images. The matching of bounding boxes, or their image patches, between images is in a non-trivial problem, therefore we treat it as a black-box task. For simplicity, we solve the matching problem by applying the pre-trained SuperGlue model~\cite{sarlin20superglue}, which can be replaced by any matching approach. From the matches we construct a view graph, two images are considered connected if they share at least five matched ROIs. In addition, only images that are connected to at least another image are used in the graph, and the graph is built so that its connections form a chain that passes through all $N$ images. This is necessary to ensure that the set of images and their matches constitute a solvable problem, in which no image view is completely disconnected from the others.

For each connected pair of images $I_i, I_j$, we estimate the relative transformation mapping from one image to the other, \ie $\hat {R}_{i,j}$ and translation $\hat {t}_{i,j}$ that satisfy: $\hat {R}_{i,j} = R_j R^T_i$ and $\hat {t}_{i,j} = t_j - R^T_j t_i$. 
We compute the relative transformation using the 5-point algorithm and RANSAC. We experiment using both BB centers, which are sensitive to the quality of the BB detection, and SuperGlue keypoint matches within the region,  which are sensitive to mismatches. 

\subsection{PoserNet: Graph-Based Relative Camera Pose Refinement}\label{sec:poser}
We define the PoserNet Graph Neural Network (GNN) as a special case of a multi-layer GNN, with shared edges representing the relative transformations and a partly shared node encoding. Therefore, we first review a general formulation of GNNs and then we describe  our  directed GNN with matched detections.

\noindent\textbf{General Graph Neural Networks:} The standard GNN formulation employs a graph $\mathcal{G} = \left \langle N,E \right \rangle$, representing a structure of nodes $N=\{N_1 \dots N_p\}$  and edges $E=\{E_1 \dots E_q\}$. Each $N_{i}$ or $E_{i}$ has an embedding ($h_i^N$,$h_i^E$) representing its latent information, which can be initialised based on \textit{a priori} information (\eg relative poses). The representations are then updated independently via a combination of message passing, nonlinear update functions ($\Psi^N(\cdot)$ and  $\Psi^E(\cdot)$) and aggregation functions. Each update constitutes a step ($k$) of the graph. The message $m^N$ between nodes from $j$ to $i$, for step ($k$), is defined as:
\begin{equation}
    m_{i,j,k}^N = \Psi^{N}(h_{i,k}^{N}, h_{j,k}^{N}, h_{i,j,k}^{E}).
\end{equation}
To update $N_i$, a message is computed for all its neighbouring nodes $j \in \mathcal{N}_i$. Finally, the aggregation function averages the incoming messages to produce the latent information for node $i$ for the $(k+1)^{th}$ iteration:
\begin{equation}
    h_{i,k+1}^{N} \leftarrow   \frac{1}{ \left | \mathcal{N}_i  \right |} \sum_{j \in \mathcal{N}_i}^{} m_{i,j,k}^N.
\end{equation}
Edges are updated in a similar fashion by applying a nonlinear transformation ($\Psi^{E}(\cdot)$) to information from the neighbour nodes and from their current latent space, \ie $h_{i,j,k+1}^{E} \leftarrow m_{i,j,k}^{E} = \Psi^{E}(h_{i,k}^{N}, h_{j,k}^{N}, h_{i,j,k}^E)$. The formulation can be generalised to multi-layer graphs simply by connecting nodes across layers and having a custom (or repeating) $\mathcal{G}$ topography for each layer.

\begin{figure}[t]
  \centering
    \includegraphics[width=1.0\columnwidth]{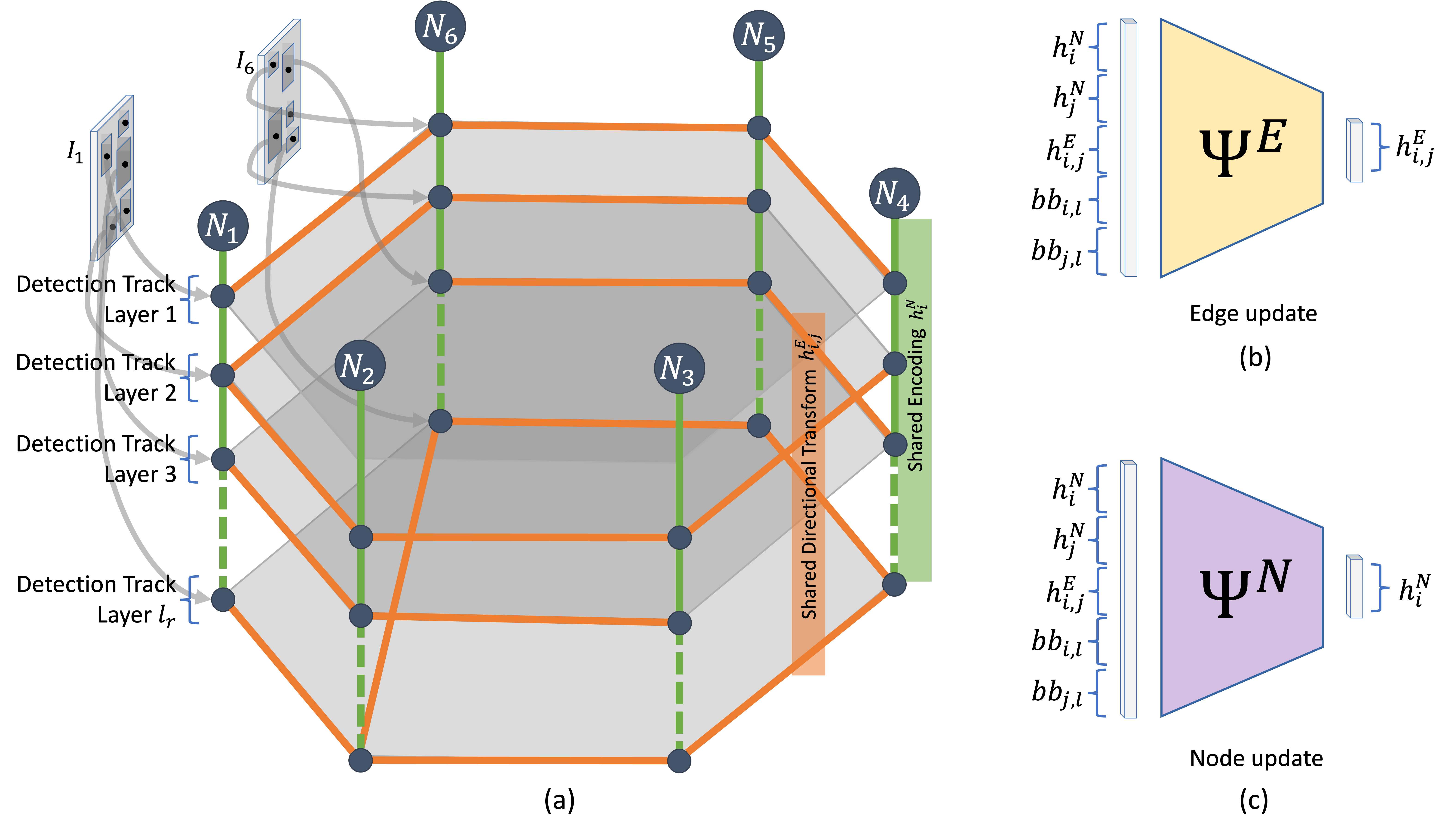}
  \caption{(a) The structure of graphs processed by PoserNet. Each image $I_i$ is associated with a node $N_{i}$ where the corresponding edges are relative transformations. Each set of corresponding detections (orange lines) between nodes refers to a layer of the graph for message passing (Detection track). (b) Nodes and (c) edges are updated by passing through $\Psi(\cdot)$ the node and edge information and respective bounding boxes ($bb$) for a given layer.
  }
  \label{fig:posernet_graph_structure}
\end{figure}

\noindent\textbf{PoserNet:} The Pose Refinement network (PoserNet)  takes as input a graph which contains the noisy relative camera poses obtained via pairwise geometry estimation (Sec.~\ref{sec:det_mat}) to initialise its edges, and it outputs a graph with the same topology, but having refined estimates for the same relative poses. Therefore,  PoserNet is a special case of the multi-layer GNN paradigm. At its simplest, it works on a graph representing an enriched version of the view graph of a scene, however, across layers of matched detections $l=\{l_1 \dots l_r\}$, with a shared node encoding ($h^{N}$), and with a shared edge representation ($h^{E}$).
See Fig.~\ref{fig:posernet_graph_structure} for a graphical representation of the structure of such a graph.

Specifically, PoserNet works on a graph 
in which nodes ($N_i$) represent camera poses, while directed edges ($E_{i,j}$) represent pairwise relative camera poses. Each node is associated with one of the input images, $I_i$. It contains the corresponding intrinsic camera parameters and it is associated with the list of bounding boxes detected on that image, arranged as layers. The embedding for a detection consists of the normalised information on the BB location, height, and width: $bb_{i,l} \in \mathbb{R}^{4}$. Detections from different nodes are connected pairwise when matched (\ie they represent the same object in multi-view), where
matched detections are represented as logical layers in the graph. Edge $E_{i,j}$ exists in the graph only if nodes $N_i$ and $N_j$ have at least one detection in common. For the message passing, the $\Psi$ functions are implemented as a multi-layer perception (MLP).

\noindent\textbf{Edge representation \& update:} The relative camera transformations are encoded in the edges as a translation vector plus a quaternion: $h^E_{i,j} \in \mathbb{R}^{7}$. We encode the transformation induced by traversing the edge in one, specified, direction, and we enforce that the transformation observed by traversing the edge in the opposite direction is the inverse of that transformation. Updating the edge representation requires creating one message per active layer ($l \in \mathcal{N}_l$), based on a pair of matched detections at its sides.
Thus, for PoserNet the message is enriched in contrast to traditional GNNs, as it takes as additional input  the representations of a pair of detections. A message is defined as: 
\begin{equation}
    m_{i,j,k,l}^{E} = \Psi^{E}(h_{i,k}^{N}, h_{j,k}^{N}, h_{i,j,k}^E, bb_{i,l}, bb_{j,l}),
\end{equation}
and the updated representation for the edge is computed as the average of the messages for all layers: $h_{i,j,k+1}^{E} \leftarrow  \frac{1}{ \left | \mathcal{N}_l  \right |} \sum_{l \in \mathcal{N}_l}^{} m_{i,j,k,l}^E.$

\noindent\textbf{Node representation \& update:} The information contained in each node consists of image height and width, the camera focal length and the first coefficient of radial distortion, all normalised: $h_{i}^{n} \in \mathbb{R}^{4}$. For updating the latent representation of a node in PoserNet, each of its neighbours can send not just one, but a set of messages, each of which corresponds to a layer with a matched pair of detections. A message is defined as:
\begin{equation}
    m_{i,j,k,l}^N = \Psi^{N}(h_{i,k}^{N}, h_{j,k}^{N}, h_{i,j,k}^{E},bb_{i,l}, bb_{j,l}),
\end{equation}
and the updated representation for the node is computed as the average of the messages from all of its neighbours ($j \in \mathcal{N}_i$), for all layers ($l \in \mathcal{N}_l$): 
\begin{equation}
h_{i,j,k+1}^{N} \leftarrow  \frac{1}{ \left | \mathcal{N}_j  \right |} \sum_{j \in \mathcal{N}_j}^{} \frac{1}{ \left | \mathcal{N}_l  \right |} \sum_{l \in \mathcal{N}_l}^{} m_{i,j,k,l}^N.
\end{equation}

This message-passing and data processing scheme is repeated a fixed number of times (one of the hyperparameters of GNNs), and it finally produces the output edge embeddings which represent the refined estimates for inter-camera pose transformations. During this process, the embedding on the nodes and edges are updated, but the information relative to bounding boxes is kept constant.

\noindent\textbf{Loss function:} PoserNet is trained using graphs for which the relative transformations between camera poses are known. The loss function comprises four components: the first and second component drive the network to produce accurate estimates for the relative poses:
\begin{equation}
    \mathcal{L}_{orient} = \angle (q_{GT}^*\circ q_{est}),
\end{equation}
\begin{equation}
        \mathcal{L}_{tr\_dir} = \angle (tr_{GT}, tr_{est}).
\end{equation}
The orientation loss, $\mathcal{L}_{orient}$, encodes the angle between ($\angle$) the ground-truth and the estimated quaternion, where $\circ$ represents quaternion composition, and $q_{GT}^*$ is the conjugate of the ground-truth quaternion. The translation direction loss, $\mathcal{L}_{tr\_dir}$, encodes the angle between ground-truth ($tr_{GT}$) and estimated translation vectors.
The remaining two components are the quaternion normalisation loss, $\mathcal{L}_{q||}$, and the translation normalisation loss, $\mathcal{L}_{tr||}$. They push the network towards producing unit norm outputs:
\begin{equation}
    \mathcal{L}_{q||}  =  \lvert \Vert q_{est}\Vert - 1 \rvert,\\
\end{equation}
\begin{equation}
    \mathcal{L}_{tr||} =  \lvert \Vert tr_{est}\Vert - 1 \rvert.\\
\end{equation}
The total loss function of PoserNet is defined as: 
\begin{equation}
    \mathcal{L}_{PoserNet} = \mathcal{L}_{orient} + \mathcal{L}_{tr\_dir} + \alpha(\mathcal{L}_{q||} + \mathcal{L}_{tr||}),
\end{equation}
where $\alpha$ is a coefficient used to tune the strength of the contribution of the different components of the loss.

\subsection{Absolute Pose Estimation}\label{motionaveragingexplained}
The relative poses refined by PoserNet can be used to compute absolute camera poses; for this, we look at State-of-the-Art methods for two common approaches: optimisation and deep learning. With the former, we want to show how modern convolution techniques can improve ``classical'' approaches, while combining PoserNet with a deep-learning model could lead to an end-to-end pipeline.

\noindent\textbf{EIG-SE3\cite{arrigoni2016spectral}:}
This method solves the Motion Averaging problem in closed-form
using spectral decomposition. Given
a fully connected graph with $n$ nodes, and accurate estimates of the
relative and absolute transformations between the cameras expressed as:
\begin{equation}\label{4x4proj}
M_{i,j}=\left({ \begin{array}{cc} R_{i,j} & t_{i,j} \\ 0 & 1 \\ \end{array} } \right),
M_{i}=\left({ \begin{array}{cc} R_{i} & t_{i} \\ 0 & 1 \\ \end{array} } \right),
\end{equation}
with $M_{i,j}\in SE(3)$, $SE(3)$ the Special Euclidean Group (3), the absolute camera poses $M_{i}\in SE(3)$, with $M_{i,j} = M_{i}M_{j}^{-1}$, can be recovered 
solving the problem $\mathbf{X} \mathbf{M} = n \mathbf{M}$, where $\mathbf{M}$ is the $4n \times 4$ matrix obtained concatenating all matrices $M_i$ and $\mathbf{X}$ is a block matrix with block $i,j = M_{i,j}$.

If the relative transformations are exact, this problem can be solved
in closed form in two steps: \emph{i)} the columns of $M$ are four eigenvectors of $\mathbf{X}$ associated with the eigenvalue $n$, found as the 4-dimensional basis of the null-space of $L=(n\mathbf{I}_{4n}-\mathbf{X})$; \emph{ii)} the basis found at the previous point is not guaranteed to be composed of Euclidean motions, \ie it is not guaranteed that each $4 \times 4$ block belongs to SE(3). This is addressed by extracting every fourth row of $M$ and finding the change of basis that maps them to $[0,0,0,1]$.

In the original work, the authors show how this approach can be generalised to problems with noisy relative transformation, by solving the Least Squares problem $\min_{M\in SE(3)} || L M ||^2_F$.
They also extend the approach to work on graphs that are not fully connected - \ie some of the off-diagonal blocks of $\mathbf{X}$ are zero - and show how the optimisation problem can be included in an iteratively reweighted Least Squares scheme to reduce the impact of outliers, which are a common occurrence in most camera pose estimation applications. See~\cite{arrigoni2016spectral} for a full description of the optimisation algorithm.

\noindent\textbf{MultiReg:} computes incremental pose updates within tangent space to ensure that poses remain on the SO(3) or SE(3) manifold. Like PoserNet, a GNN is used where, however,  nodes represent camera poses and the edges represent transformations, $\mathcal{G}_{a} = \left \langle N_a,E_a \right \rangle$. In addition,  only the absolute pose and a latent representation is updated on a node, with no edge update. To achieve this each increment $\epsilon_i$ is computed from the pose residuals $\Gamma_{i,j}=T_{i}T_{j}^{-1}\hat{T}_{i,j}^{-1}$ based on the current absolute pose and the latent encoding ($h^N$) stored in the node. Therefore each update is equivalent to:
\begin{equation}
    \left ( \epsilon_i,\Delta h_i^N \right )\leftarrow \Psi^N (h_i^N,u,\Psi^E(h_i^N,h_j^N\Gamma_{i,j})),
\end{equation}
where $\Psi^N$ and $\Psi^E$ are MLPs and $u$ is a global encoding from average pooling of $f$ over the graph. The transform update is therefore $T_{i} \leftarrow \text{exp}(\hat{\epsilon}_{i}) T_i$ which corresponds to addition in Euclidean space for Lie Groups. Finally, similar to pair-wise we compute the loss based on the decomposed rotation and translation.
\begin{equation}
    \mathcal{L}_{abs} = \frac{1}{\left | \epsilon^{c}  \right |}\sum_{(i,j)\in\epsilon^{c}}\left ( \left | R_{i,j} - R_{i,j}^{GT} \right | + \lambda_{t} \left | t_{i,j} - t_{i,j}^{GT} \right | \right ).
\end{equation}

\section{Experiments}
In this section, we provide an evaluation of the proposed PoserNet
model. We validate our choice of PoserNet structure (\ie depth, choice of training data), and show how
PoserNet can improve the accuracy of relative pose estimation. We then investigate the effects of this refinement on Motion Averaging.
As an implementation detail, PoserNet partially relies on PyTorch Geometric (PyG)~\cite{Fey/Lenssen/2019}.

For each experiment we report separately the error on the camera rotation and the translation error; the former is expressed in degrees, 
while the translation is reported in meters for absolute poses and in degrees for relative poses. This is due to the fact that relative poses are defined up to a scale, and we therefore report the angular distance between the ground-truth and the estimated translation unit vector.
These results are expressed in form of the median error $\eta$ (the smaller the better) and
as the percentage of testing graphs with and average error smaller than predefined thresholds (the higher the better).
Such thresholds are $3$, $5$, $10$, $30$ and $45$ degrees for the angular metrics, and  $0.05$, $0.1$, $0.25$, $0.5$ and $0.75$ for errors expressed in meters. 
For all experiments that required training network, we trained three separate models and report their average performance, with their standard deviation. 

\subsection{Dataset}
For our experiments, we use graphs built from the 7-Scenes~\cite{glocker2013real-time}: this is a collection
of $41k$ images acquired in seven small-sized indoor environments with a Microsoft Kinect. The RGB-D information has been used to compute accurate ground-truth camera poses. 
For each scene, several sequences were recorded by different users, and they are split into distinct training and testing sequence sets. Starting from 7-Scenes data, we created two datasets comprising graphs of different sizes: a small graphs dataset and a larger graph dataset. In each case, we followed the split between training and test data or the original work.

\noindent\textbf{Small graphs dataset:} It contains $8$-node graphs. For each scene we created $2k$ graphs from the training sequences, and $1k$ from the test ones. Each graph was created by randomly selecting eight images that satisfy the connectivity constraints outlined in Section~\ref{sec:det_mat}. 
We thus ensure that all graph nodes are connected, while not requiring the graph to be fully connected. 

\noindent\textbf{Large graphs dataset:} The graphs are generated sampling, 
for each scene, $125$ frames from the aggregated training or testing sequences. In this case, we do not explicitly enforce a connectivity
constraint, because the limited size of the scenes and the large number of nodes guarantee a large number of connected edges. We extracted only one training and one testing graphs from the first five scenes, due to the difficulty in identifying large graphs.

\begin{table}[t]
\begin{center}
\scriptsize
\begin{tabularx}{\textwidth}{X c c c c c c c}
 \toprule
  Dataset part & Graphs & Nodes & Edges & Nodes/graph & Edges/graph & Edges/node \\ 
  \midrule
 Small graphs, train  & 14000  & 112000  & 215790  &   8  &   15.4  & 1.9  \\ 
 Small graphs, test   &  7000  &  56000  & 110367  &   8  &   15.8  & 2.0  \\ 
 Large graphs, train    &     5  &    625  &  22674  & 125  & 4534.8  & 36.3 \\ 
 Large graphs, test     &     5  &    625  &  22606  & 125  & 4521.2  & 36.2 \\ 
 \bottomrule
\end{tabularx}
\end{center}
\caption{\label{tab:datasets} Composition of the small and large graph datasets used to validate the proposed method. The different graph-building strategies result in large graphs which have a higher (18-fold higher) connectivity than the small graphs.}
\end{table}

\subsection{PoserNet Refining Relative Poses}
We present an analysis designed to assess the effectiveness of PoserNet in refining relative camera poses. 
We first show how PoserNet successfully improves relative poses provided by different initialisation algorithms, and secondly how PoserNet generalises well to graphs of different sizes. In Supplementary Material we discuss the impact of choosing different depths for PoserNet's GNN, we identify a depth of $2$ as the best and we use it throughout the remaining experiments. 

\subsubsection{Refining Relative Poses from Different Initialisation Modes:} We assessed the effectiveness of PoserNet in 
refining the relative pose estimates initialised from bounding box centres (BB) and keypoints (KP), as discussed in Section~\ref{sec:det_mat}. We evaluate the PoserNet model on the small graphs dataset as shown in Table ~\ref{tab:results_posernet_bb_vs_keypoints}.
In both cases we see a significant performance gain, with median rotation error reduction of over $76\degree$ and $28\degree$ for the BB and KP cases respectively, and an improvement of over $42\degree$ and $72\degree$ for the translation median angular error.
This confirms PoserNet's effectiveness at refining relative poses generated by different algorithms, with different levels of noise. 

\begin{table*}[htp!]
\begin{minipage}{1.0\textwidth}
    \scriptsize
    \centering
    \begin{tabularx}{\textwidth}{k k c c c c c c c c c c c c}
    \toprule
         &         & \multicolumn{6}{c}{Orientation error (deg)}        & \multicolumn{6}{c}{Translation direction error (deg)}               \\ \cmidrule(lr){3-8} \cmidrule(lr){9-14}
    Input   &  & 3 & 5 & 10 & 30 & 45 & $\eta$ & 3 & 5 & 10 & 30 & 45 & $\eta$ \\\midrule
    \multirow{3}{*}{BB} & Initial      & 
       0.00 & 0.00 &0.00 &0.16 &1.44 &96.48 &0.00 &0.00 &0.00 &0.00 &0.00 &89.30
    \\  \cmidrule(lr){3-14}
         & PoserNet      & 
         \begin{tabular}[c]{@{}c@{}}0.00\\ {\tiny$\pm$0.00}\end{tabular}  & 
         \begin{tabular}[c]{@{}c@{}}0.00\\{\tiny$\pm$0.00}\end{tabular}   &  
         \begin{tabular}[c]{@{}c@{}}0.74\\{\tiny$\pm$0.14}\end{tabular}   & 
         \begin{tabular}[c]{@{}c@{}}88.18\\{\tiny$\pm$0.28}\end{tabular}  & 
         \begin{tabular}[c]{@{}c@{}}99.02\\{\tiny$\pm$0.02}\end{tabular}  &       
         \begin{tabular}[c]{@{}c@{}}20.39\\{\tiny$\pm$0.20}\end{tabular}  &
         
         \begin{tabular}[c]{@{}c@{}}0.00\\{\tiny$\pm$0.00}\end{tabular}   & 
         \begin{tabular}[c]{@{}c@{}}0.00\\{\tiny$\pm$0.00}\end{tabular}   &   
         \begin{tabular}[c]{@{}c@{}}0.00\\{\tiny$\pm$0.00}\end{tabular}   & 
         \begin{tabular}[c]{@{}c@{}}7.92\\{\tiny$\pm$0.25}\end{tabular}   &  
         \begin{tabular}[c]{@{}c@{}}45.63\\{\tiny$\pm$0.43}\end{tabular}  &       
         \begin{tabular}[c]{@{}c@{}}46.60\\{\tiny$\pm$0.14}\end{tabular} \\
         \midrule

    \multirow{3}{*}{KP}                                                        & 
        Initial      &  
        0.74 & 4.01 &11.16 &41.46 &62.53 &36.26 &0.00 &0.03 &0.10 &1.37 &5.49 &87.23
             
        \\ \cmidrule(lr){3-14}
         & PoserNet      & 
         \begin{tabular}[c]{@{}c@{}}1.13\\ {\tiny$\pm$0.11}\end{tabular} & 
         \begin{tabular}[c]{@{}c@{}}17.24\\{\tiny$\pm$0.86}\end{tabular} &
         \begin{tabular}[c]{@{}c@{}}78.19\\{\tiny$\pm$1.36}\end{tabular} & 
         \begin{tabular}[c]{@{}c@{}}98.78\\{\tiny$\pm$0.02}\end{tabular} & 
         \begin{tabular}[c]{@{}c@{}}99.86\\{\tiny$\pm$0.01}\end{tabular} &
         \begin{tabular}[c]{@{}c@{}}7.31\\{\tiny$\pm$0.11}\end{tabular}  &
         
         \begin{tabular}[c]{@{}c@{}}0.01\\{\tiny$\pm$0.01}\end{tabular}  &
         \begin{tabular}[c]{@{}c@{}}1.42\\{\tiny$\pm$0.28}\end{tabular}  & 
         \begin{tabular}[c]{@{}c@{}}27.67\\{\tiny$\pm$0.70}\end{tabular} & 
         \begin{tabular}[c]{@{}c@{}}80.07\\{\tiny$\pm$0.31}\end{tabular} &  
         \begin{tabular}[c]{@{}c@{}}90.01\\{\tiny$\pm$0.68}\end{tabular} &  
         \begin{tabular}[c]{@{}c@{}}14.54\\{\tiny$\pm$0.14}\end{tabular} \\ 
    \bottomrule
    \end{tabularx}
    \centering
    \caption{\label{tab:results_posernet_bb_vs_keypoints}
    PoserNet performances in refining relative pose computed from the centres of matched bounding boxes (BB) or from a standard keypoints approach (KP). For both cases, the initial orientation and translation direction error are improved by PoserNet processing.}
\end{minipage}
\begin{minipage}{1.0\textwidth}
    \scriptsize
    \centering
    \begin{tabularx}{\textwidth}{w w c c c c c c c c c c c c}
    \toprule
         &         & \multicolumn{6}{c}{Orientation error (deg)}        & \multicolumn{6}{c}{Translation direction error (deg)}               \\ \cmidrule(lr){3-8} \cmidrule(lr){9-14}
    Train & Test & 3 & 5 & 10 & 30 & 45 & $\eta$ & 3 & 5 & 10 & 30 & 45 & $\eta$ \\\midrule
    Raw & Small      &  0.74 & 4.01 &11.16 &41.46 &62.53 &36.26 &0.00 &0.03 &0.10 &1.37 &5.49 &87.23 \\
         \cmidrule(lr){3-14}
         
         Small & Small      & 
         \begin{tabular}[c]{@{}c@{}}1.13\\ {\tiny$\pm$0.11}\end{tabular} & 
         \begin{tabular}[c]{@{}c@{}}17.24\\{\tiny$\pm$0.86}\end{tabular} &
         \begin{tabular}[c]{@{}c@{}}78.19\\{\tiny$\pm$1.36}\end{tabular} & 
         \begin{tabular}[c]{@{}c@{}}98.78\\{\tiny$\pm$0.02}\end{tabular} & 
         \begin{tabular}[c]{@{}c@{}}99.86\\{\tiny$\pm$0.01}\end{tabular} &
         \begin{tabular}[c]{@{}c@{}}\textbf{7.31}\\{\tiny$\pm$0.11}\end{tabular}  &
         
         \begin{tabular}[c]{@{}c@{}}0.01\\{\tiny$\pm$0.01}\end{tabular}  &
         \begin{tabular}[c]{@{}c@{}}1.42\\{\tiny$\pm$0.28}\end{tabular}  & 
         \begin{tabular}[c]{@{}c@{}}27.67\\{\tiny$\pm$0.70}\end{tabular} & 
         \begin{tabular}[c]{@{}c@{}}80.07\\{\tiny$\pm$0.31}\end{tabular} &  
         \begin{tabular}[c]{@{}c@{}}90.01\\{\tiny$\pm$0.68}\end{tabular} &  
         \begin{tabular}[c]{@{}c@{}}\textbf{14.54}\\{\tiny$\pm$0.14}\end{tabular} \\ 
         \cmidrule(lr){3-14}
         
         Large & Small      & 
         \begin{tabular}[c]{@{}c@{}}0.00\\ {\tiny$\pm$0.00}\end{tabular} & 
         \begin{tabular}[c]{@{}c@{}}0.04\\{\tiny$\pm$0.03}\end{tabular}  &
         \begin{tabular}[c]{@{}c@{}}13.26\\{\tiny$\pm$5.55}\end{tabular} & 
         \begin{tabular}[c]{@{}c@{}}96.98\\{\tiny$\pm$0.37}\end{tabular} & 
         \begin{tabular}[c]{@{}c@{}}99.61\\{\tiny$\pm$0.09}\end{tabular} &
         \begin{tabular}[c]{@{}c@{}}13.97\\{\tiny$\pm$0.94}\end{tabular} &
         
         \begin{tabular}[c]{@{}c@{}}0.00\\{\tiny$\pm$0.00}\end{tabular}  &
         \begin{tabular}[c]{@{}c@{}}0.00\\{\tiny$\pm$0.00}\end{tabular}  & 
         \begin{tabular}[c]{@{}c@{}}0.11\\{\tiny$\pm$0.10}\end{tabular}  & 
         \begin{tabular}[c]{@{}c@{}}57.19\\{\tiny$\pm$5.72}\end{tabular} &  
         \begin{tabular}[c]{@{}c@{}}81.26\\{\tiny$\pm$1.16}\end{tabular} &  
         \begin{tabular}[c]{@{}c@{}}27.56\\{\tiny$\pm$2.02}\end{tabular} \\ 
         \midrule

        Raw & Large      &  0.00 & 0.00 &0.00 &20.00 &60.00 &37.70 &0.00 &0.00 &0.00 &0.00 &0.00 &79.60\\
         \cmidrule(lr){3-14}
         
        Small & Large       &  
         \begin{tabular}[c]{@{}c@{}}0.00\\ {\tiny$\pm$0.00}\end{tabular}  & 
         \begin{tabular}[c]{@{}c@{}}0.00\\{\tiny$\pm$0.00}\end{tabular}   &
         \begin{tabular}[c]{@{}c@{}}80.0\\{\tiny$\pm$0.00}\end{tabular}   & 
         \begin{tabular}[c]{@{}c@{}}100.00\\{\tiny$\pm$0.00}\end{tabular} & 
         \begin{tabular}[c]{@{}c@{}}100.00\\{\tiny$\pm$0.00}\end{tabular} &
         \begin{tabular}[c]{@{}c@{}}\textbf{6.85}\\{\tiny$\pm$0.11}\end{tabular}   &
         
         \begin{tabular}[c]{@{}c@{}}0.00\\{\tiny$\pm$0.00}\end{tabular}   &
         \begin{tabular}[c]{@{}c@{}}0.00\\{\tiny$\pm$0.00}\end{tabular}   & 
         \begin{tabular}[c]{@{}c@{}}20.00\\{\tiny$\pm$0.00}\end{tabular}  & 
         \begin{tabular}[c]{@{}c@{}}100.00\\{\tiny$\pm$0.00}\end{tabular} &  
         \begin{tabular}[c]{@{}c@{}}100.00\\{\tiny$\pm$0.00}\end{tabular} &  
         \begin{tabular}[c]{@{}c@{}}\textbf{12.62}\\{\tiny$\pm$0.48}\end{tabular}  \\
         \cmidrule(lr){3-14}
         
         Large & Large      & 
         \begin{tabular}[c]{@{}c@{}}0.00\\ {\tiny$\pm$0.00}\end{tabular}  & 
         \begin{tabular}[c]{@{}c@{}}0.00\\{\tiny$\pm$0.00}\end{tabular}   &
         \begin{tabular}[c]{@{}c@{}}13.33\\{\tiny$\pm$9.43}\end{tabular}  & 
         \begin{tabular}[c]{@{}c@{}}100.00\\{\tiny$\pm$0.00}\end{tabular} & 
         \begin{tabular}[c]{@{}c@{}}100.00\\{\tiny$\pm$0.00}\end{tabular} &
         \begin{tabular}[c]{@{}c@{}}12.11\\{\tiny$\pm$1.10}\end{tabular}  &
         
         \begin{tabular}[c]{@{}c@{}}0.00\\{\tiny$\pm$0.00}\end{tabular}   &
         \begin{tabular}[c]{@{}c@{}}0.00\\{\tiny$\pm$0.00}\end{tabular}   & 
         \begin{tabular}[c]{@{}c@{}}0.00\\{\tiny$\pm$0.00}\end{tabular}   & 
         \begin{tabular}[c]{@{}c@{}}100.00\\{\tiny$\pm$0.00}\end{tabular} &  
         \begin{tabular}[c]{@{}c@{}}100.00\\{\tiny$\pm$0.00}\end{tabular} &  
         \begin{tabular}[c]{@{}c@{}}21.58\\{\tiny$\pm$1.36}\end{tabular}  \\ 
         \bottomrule
         
    \end{tabularx}
    \centering
    \caption{\label{tab:results_posernet_small_vs_large_graphs}
    PoserNet performances when trained and evaluated on small (Small) or large (Large) graphs. While versions of PoserNet trained on either largely outperform the baseline error (Raw), models trained on small give the best performances on both testing datasets
    (Highlighted in bold).}
\end{minipage}

\begin{minipage}{1.0\textwidth}
    \scriptsize
    \centering
    \begin{tabularx}{\textwidth}{w w c@{\hskip 0.06in} c@{\hskip 0.06in} c@{\hskip 0.06in} c@{\hskip 0.06in} c@{\hskip 0.06in} c@{\hskip 0.06in} c@{\hskip 0.06in} c@{\hskip 0.06in} c@{\hskip 0.06in} c@{\hskip 0.06in} c@{\hskip 0.06in} c@{\hskip 0.06in} c}

    \toprule
         &         & \multicolumn{6}{c}{Orientation error (deg)}        & \multicolumn{6}{c}{Translation error (meters)}               \\ \cmidrule(lr){3-8} \cmidrule(lr){10-15}
    Model                                                                            & Init & 3 & 5 & 10 & 30 & 45 & $\eta$ & & 0.05 & 0.1 & 0.25 & 0.5 & 0.75 & $\eta$ \\\midrule
    \multirow{3}{*}{MultiReg}                                                        & BB      &  \begin{tabular}[c]{@{}c@{}}0.60\\ {\tiny$\pm$0.00}\end{tabular}  & \begin{tabular}[c]{@{}c@{}}2.10\\{\tiny$\pm$0.00}\end{tabular}  &  \begin{tabular}[c]{@{}c@{}}8,10\\{\tiny$\pm$0.10}\end{tabular}  & \begin{tabular}[c]{@{}c@{}}47.60\\{\tiny$\pm$0.53}\end{tabular}   & \begin{tabular}[c]{@{}c@{}}69.13\\{\tiny$\pm$0.47}\end{tabular}   &       \begin{tabular}[c]{@{}c@{}}31.37\\{\tiny$\pm$0.30}\end{tabular}   & &  \begin{tabular}[c]{@{}c@{}}0.50\\{\tiny$\pm$0.00}\end{tabular}    &  \begin{tabular}[c]{@{}c@{}}1.47\\{\tiny$\pm$0.06}\end{tabular}   &   \begin{tabular}[c]{@{}c@{}}6.93\\{\tiny$\pm$0.12}\end{tabular}   &  \begin{tabular}[c]{@{}c@{}}23.03\\{\tiny$\pm$0.70}\end{tabular}   &   \begin{tabular}[c]{@{}c@{}}40.80\\{\tiny$\pm$0.85}\end{tabular}  &       \begin{tabular}[c]{@{}c@{}}0.89\\{\tiny$\pm$0.01}\end{tabular}             
    \\  \cmidrule(lr){3-15}
         & KP      & \begin{tabular}[c]{@{}c@{}}10.23\\ {\tiny$\pm$0.12}\end{tabular}  & \begin{tabular}[c]{@{}c@{}}24.23\\{\tiny$\pm$0.25}\end{tabular}  &  \begin{tabular}[c]{@{}c@{}}39.7\\{\tiny$\pm$0.40}\end{tabular}  & \begin{tabular}[c]{@{}c@{}}67.97\\{\tiny$\pm$0.06}\end{tabular}   & \begin{tabular}[c]{@{}c@{}}79.90\\{\tiny$\pm$0.00}\end{tabular}   &       \begin{tabular}[c]{@{}c@{}}\bf{15.69}\\{\tiny$\pm$0.08}\end{tabular}  & & \begin{tabular}[c]{@{}c@{}}1.00\\{\tiny$\pm$0.00}\end{tabular}    &  \begin{tabular}[c]{@{}c@{}}3.77\\{\tiny$\pm$0.06}\end{tabular}   &   \begin{tabular}[c]{@{}c@{}}17.77\\{\tiny$\pm$0.12}\end{tabular}   &  \begin{tabular}[c]{@{}c@{}}40.30\\{\tiny$\pm$0.17}\end{tabular}   &   \begin{tabular}[c]{@{}c@{}}57.23\\{\tiny$\pm$0.06}\end{tabular}  &       \begin{tabular}[c]{@{}c@{}}\bf{0.64}\\{\tiny$\pm$0.01}\end{tabular}\\
         \midrule

    \multirow{3}{*}{\begin{tabular}[c]{@{}l@{}}MultiReg\\ + \\ PoserNet\end{tabular}}                                                        & BB      &  \begin{tabular}[c]{@{}c@{}}0.50\\ {\tiny$\pm$0.00}\end{tabular}  & \begin{tabular}[c]{@{}c@{}}1.60\\{\tiny$\pm$0.00}\end{tabular}  &  \begin{tabular}[c]{@{}c@{}}6.10\\{\tiny$\pm$0.10}\end{tabular}  & \begin{tabular}[c]{@{}c@{}}41.03\\{\tiny$\pm$0.42}\end{tabular}   & \begin{tabular}[c]{@{}c@{}}62.50\\{\tiny$\pm$0.14}\end{tabular}   &       \begin{tabular}[c]{@{}c@{}}35.60\\{\tiny$\pm$0.33}\end{tabular}   & & \begin{tabular}[c]{@{}c@{}}0.60\\{\tiny$\pm$0.00}\end{tabular}    &  \begin{tabular}[c]{@{}c@{}}1.50\\{\tiny$\pm$0.00}\end{tabular}   &   \begin{tabular}[c]{@{}c@{}}6.53\\{\tiny$\pm$0.06}\end{tabular}   &  \begin{tabular}[c]{@{}c@{}}21.60\\{\tiny$\pm$0.00}\end{tabular}   &   \begin{tabular}[c]{@{}c@{}}39.33\\{\tiny$\pm$0.06}\end{tabular}  &       \begin{tabular}[c]{@{}c@{}}0.91\\{\tiny$\pm$0.00}\end{tabular}             
    \\ \cmidrule(lr){3-15}
         & KP      & \begin{tabular}[c]{@{}c@{}}0.50\\ {\tiny$\pm$0.00}\end{tabular}  & \begin{tabular}[c]{@{}c@{}}1.63\\{\tiny$\pm$0.06}\end{tabular}  &  \begin{tabular}[c]{@{}c@{}}6.20\\{\tiny$\pm$0.10}\end{tabular}  & \begin{tabular}[c]{@{}c@{}}40.90\\{\tiny$\pm$0.44}\end{tabular}   & \begin{tabular}[c]{@{}c@{}}62.60\\{\tiny$\pm$0.53}\end{tabular}   &       \begin{tabular}[c]{@{}c@{}}35.73\\{\tiny$\pm$0.37}\end{tabular}  & &  \begin{tabular}[c]{@{}c@{}}0.60\\{\tiny$\pm$0.00}\end{tabular}    &  \begin{tabular}[c]{@{}c@{}}1.50\\{\tiny$\pm$0.00}\end{tabular}   &   \begin{tabular}[c]{@{}c@{}}6.40\\{\tiny$\pm$0.10}\end{tabular}   &  \begin{tabular}[c]{@{}c@{}}21.43\\{\tiny$\pm$0.12}\end{tabular}   &   \begin{tabular}[c]{@{}c@{}}39.23\\{\tiny$\pm$0.12}\end{tabular} &       \begin{tabular}[c]{@{}c@{}}0.91\\{\tiny$\pm$0.00}\end{tabular} \\
                 \midrule
                 \midrule

    \multirow{2}{*}{\begin{tabular}[c]{@{}l@{}}EIG-SE3 \end{tabular}} & BB & 
    	0.00 & 0.00 & 0.09 & 2.46 & 7.84 & 78.46 & & 0.00 & 0.00 & 0.00 & 0.57 & 13.47 & 0.65
    \\  \cmidrule(lr){3-15}
    & KP & 
    11.99 & 23.87 & 34.49 & 57.59 & 62.09 & 25.37 & & 0.00 & 0.06 & 2.63 & 31.39 & 74.40 & 0.61 \\
    \midrule
    \multirow{2}{*}{\begin{tabular}[c]{@{}l@{}}EIG-SE3\\ +\\ PoserNet \end{tabular}} & BB &
	0.00 & 0.09 & 10.36 & 88.00 & 94.06 & 15.80 & & 0.00 & 0.00 & 7.54 & 46.80 & 83.23 & 0.52
    \\  \cmidrule(lr){3-15}
     & KP  &
	15.56 & 48.36 & 82.14 & 93.14 & \hphantom 95.69 & \bf{5.10} & & 0.04 & 1.94 & 34.51 & 76.43 & 94.49 & \bf{0.32} \\
    \addlinespace[1ex]
	\bottomrule
    \end{tabularx}
    \centering
    \caption{\label{tab:results_7scenes} Motion averaging performances on the 7-Scenes dataset using the MultiReg 
    and EIG-SE3 algorithms. We report results for  different initialisation methods: bounding box centres (BB) and keypoints (KP).}
\end{minipage}
\end{table*}

\subsubsection{Dealing with graphs of different sizes:} We explore the impact of using the large and small graphs to train and test PoserNet. The results are summarised in Table~\ref{tab:results_posernet_small_vs_large_graphs}. For each test set, we provide the baseline errors of the unrefined graphs, and the performances of models trained on either the large graph or small graph dataset.
From the results, we make three observations. First, we verify that PoserNet is effective at refining relative poses: models trained on either small or large graphs and tested on the corresponding test set achieve a significant accuracy improvement over the baseline. Second, we see that PoserNet has good generalisation capabilities: the pose refinement performance generalises quite well when training on one dataset, \eg small graphs, and testing on large graphs. Finally, we observe that models trained on small graphs perform better, while testing on large graphs leads to more accurate estimates. 

\subsubsection{Generalisation to unseen graphs:} To test the ability of our method to generalise to novel scenes, we evaluate PoserNet's performance on relative pose refinement for 8-node graphs using a leave-one-out scheme. This is done evaluating on each of the seven scenes of the dataset a model trained only on the other six scenes. The results of this test, reported in the Supplementary Material, show median orientation and translation orientation errors of $8.04\degree$ and $19.75\degree$ respectively, comparable with the performance on the full dataset ($7.31\degree$ and $14.54\degree$).  

\subsection{Absolute Pose Estimates}

The last experiments show the impact of PoserNet on the motion
averaging task (\ie extract absolute camera poses from the relative ones). We consider two possible approaches: a modern, graph-network based
model (MultiReg); and an optimisation-based algorithm (EIG-SE3).
Results for both Motion Averaging approaches are reported in Table~\ref{tab:results_7scenes},
using the same notation of previous experiments. However, since the Motion Averaging
process reintroduces the scale of the translation vector, the error on the translations
is reported as the Euclidean distance between the ground-truth and estimated vectors, expressed
in meters.

\subsubsection{Motion averaging via MultiReg:}
In the case of MultiReg, the refinement of the relative poses via PoserNet does not
improve the performance, leading instead to larger errors. Specifically, refining the
poses makes the orientation and translation errors respectively $4.2\degree$ and $0.02$ m
worse in the case of BB initialisation, and $20\degree$ and $0.27$ m
worse in the case of KP initialisation. 

While this performance is disappointing, it is not completely unexpected. Firstly, the original MultiReg model was tested on large, very connected graphs, which means its structure and parameters might not be well-suited for our small graph dataset. On the other hand, the MultiReg model is too complex to be trained only on our limited large graph dataset. Moreover, \cite{yew2020-RobustSync} discarded as outliers all edges initialised with a rotation error larger than $15\degree$ and a translation error of more than $15$ cm, which makes our initial error condition very challenging in comparison.

\subsubsection{Motion averaging via EIG-SE3:}
Finally, we show the performance gain obtained by combining PoserNet with a closed-form 
approach to solve Motion Averaging, the EIG-SE3. Unlike MultiReg, this approach does not
require training, making it easier to deploy and more general. Given the deterministic
nature of this model, we report results of only one evaluation run.

Results of this experiment are reported in the bottom two rows of Table~\ref{tab:results_7scenes};
in this case, PoserNet results in improvement over the baseline
error for both the $BB$ and $KP$ initialisation modes. In the $BB$ case, the median error on the rotation improves by $62\degree$, and in the $KP$ case the orientation error is reduced by $20 \degree$. Performance on the translation are also clearly noticeable, with an improvement of $0.13$ m and $0.29$ m for the BB and KP initialisation respectively.

\section{Conclusions}
We have proposed the novel PoserNet module for relative pose refinement driven by matched detections, based on ROIs generated from objectness-based detectors.
We have shown that such ROIs are sufficient to recover accurate camera poses, with a large improvement with respect to the initial pair-wise estimates. In addition, we have shown how PoserNet can improve the outcome of the optimisation-based Motion Averaging EIG-SE3. The proposed model is a relatively simple and fast network relying on sparse object detections (in contrast to keypoints) and relatively lightweight bounding box representation. Therefore, the combination with a `classical' geometric approach, such as EIG-SE3, yields increased performance compared to the more complex state-of-the-art networks like MultiReg. While the lack of improvement in combination with MultiReg highlights a challenge of deep learning methods being able to generalise to other scenarios outside of their initial design. We point to this limitation as a direction for future work, which could result in a generalised Motion Averaging method as flexible as optimisation methods and able to incorporate specific elements such as PoserNet for task or domain-specific improvement. 
\vfill

{\small
\bibliographystyle{ieee_fullname}
\bibliography{egbib}
}

\appendix
\chapter*{Supplementary material}

\section{Introduction}

In the main paper, we introduced a novel approach to relative pose refinement (PoserNet) and evaluated
its effectiveness in correcting initial relative pose estimates. We then investigated the effect of PoserNet on 
Motion Averaging performed via either an optimisation-based method (EIG-SE3~\cite{arrigoni2016spectral}), or a deep-learning-based method (MultiReg~\cite{yew2020-RobustSync}). 
In both cases, we used as training and testing dataset a collection of graphs generated using images from 7-Scenes\cite{glocker2013real-time}.

In this document, we provide additional details about
\emph{i)} how the 7-Scenes graphs were obtained by generating and matching ROI detections; 
\emph{ii)} the implementation details for our PoserNet and for the third-party MultiReg~\cite{yew2020-RobustSync} and EIG-SE3~\cite{arrigoni2016spectral} methods;
\emph{iii)} a more detailed discussion of the error distribution after motion averaging (Table\ref{tab:results_7scenes} of the main paper);
\emph{iv)} the effect of increasing PoserNet's complexity by increasing its ``depth'';
\emph{v)} PoserNet generalisation capabilities.

\section{Generating Graphs from 7-Scenes}

\begin{figure}[th!]
   \centering
   \begin{subfigure}[b]{0.56\columnwidth}
     \includegraphics[width=\columnwidth]{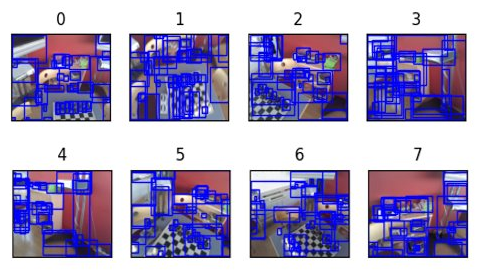}
     \caption{Objectness Detections}
   \end{subfigure}
   \begin{subfigure}[b]{0.4\columnwidth}
     \includegraphics[width=\columnwidth]{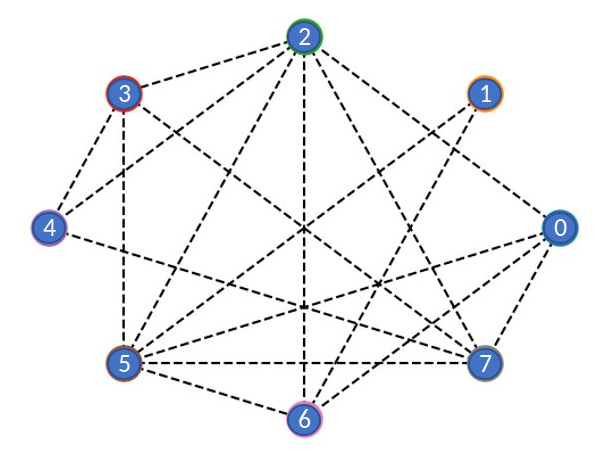}
     \caption{Sample Graph}
   \end{subfigure} \\
   \begin{subfigure}[b]{\columnwidth}
     \includegraphics[width=\columnwidth]{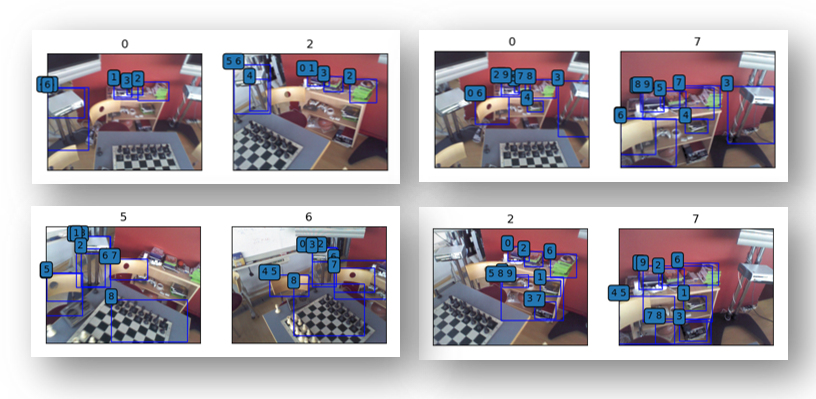}
     \caption{Matched Detections}
   \end{subfigure}

   \caption{Sample input graph. We select eight images and obtain objectness detections via OLN (a) and use them as nodes in a graph (b). The number on top of each image represents the image ID, which also corresponds to the node ID as shown in (b). Two nodes are connected if they share at least five matched detections. In (c) we show matched detections for pairs of images. Detections matched across each image pair and indicated with the same number (blue box).}
   \label{fig:sample_graph}
\end{figure}

In this section, we provide additional details on how we generated object detections and on how we established matches between them.
We provide an example of the resulting graphs in Figure~\ref{fig:sample_graph}.
Moreover, we share general considerations on the difficulties we encountered in the process.  

\begin{figure}
    \centering
    \begin{subfigure}[b]{0.4\columnwidth}
    \includegraphics[width=\columnwidth]{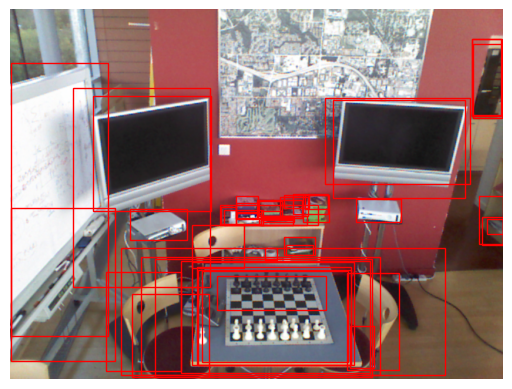}
    \caption{Chess - R-CNN}
    \end{subfigure}
    \begin{subfigure}[b]{0.4\columnwidth}
    \includegraphics[width=\columnwidth]{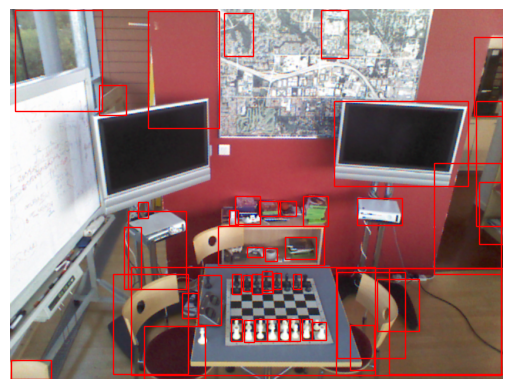}
    \caption{Chess - OLN}
    \end{subfigure}
    \begin{subfigure}[b]{0.4\columnwidth}
    \includegraphics[width=\columnwidth]{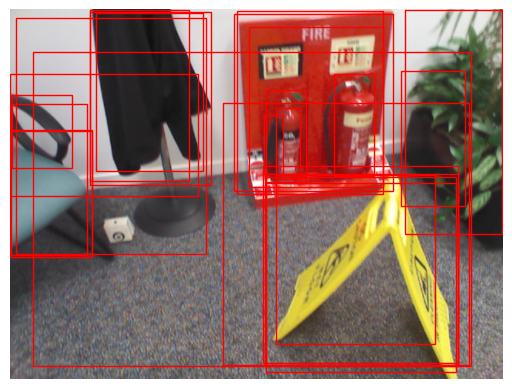}
    \caption{Fire - R-CNN}
    \end{subfigure}
    \begin{subfigure}[b]{0.4\columnwidth}
    \includegraphics[width=\columnwidth]{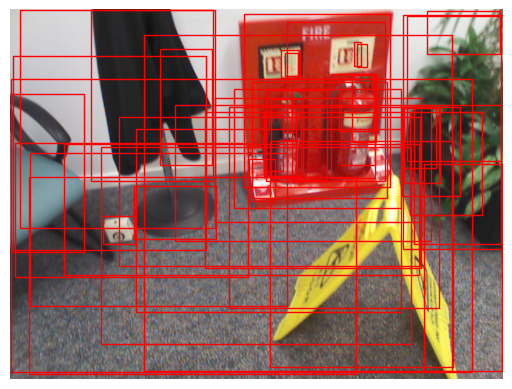}
    \caption{Fire - OLN}
    \end{subfigure}
    \begin{subfigure}[b]{0.4\columnwidth}
    \includegraphics[width=\columnwidth]{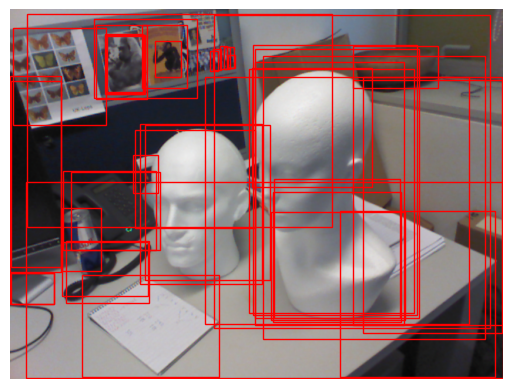}
    \caption{Heads - R-CNN}
    \end{subfigure}
    \begin{subfigure}[b]{0.4\columnwidth}
    \includegraphics[width=\columnwidth]{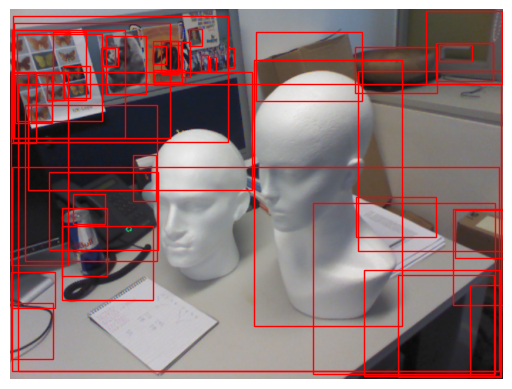}
    \caption{Heads - OLN}
    \end{subfigure}
    \begin{subfigure}[b]{0.4\columnwidth}
    \includegraphics[width=\columnwidth]{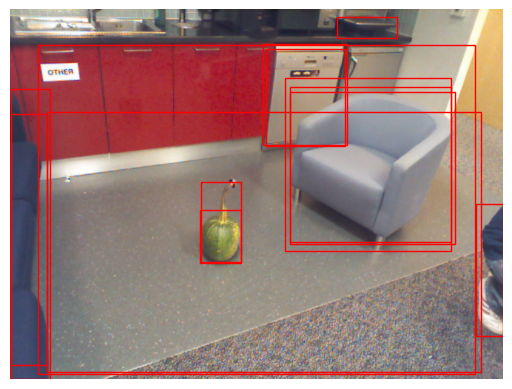}
    \caption{Pumpkin - R-CNN}
    \end{subfigure}
    \begin{subfigure}[b]{0.4\columnwidth}
    \includegraphics[width=\columnwidth]{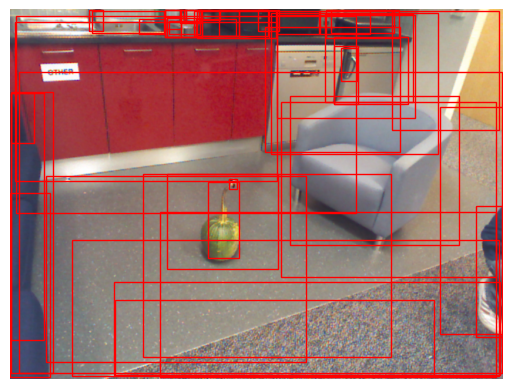}
    \caption{Pumpkin - OLN}
    \end{subfigure}
  \caption{Different approaches to ROI generation: ROIs computed with the pre-trained  ResNet50 
  component of the Faster R-CNN~\cite{NIPS2015_14bfa6bb} network, and by OLN~\cite{kim2021oln}. We found OLN  more
  reliable at identifying small and re-occurring objects.
  \label{fig:7scenes_rois}}
\end{figure}

\noindent{}\textbf{ROIs generation --} We computed ROIs for our experiment applying the pre-trained Object Localisation Network (OLN)~\cite{kim2021oln} on the images of the 7-Scenes dataset. Examples of ROIs computed via OLN, and a comparison with those obtained using a part of Faster R-CNN~\cite{NIPS2015_14bfa6bb} are shown in Figure~\ref{fig:7scenes_rois}. We opted for using OLN in our work because we found it more reliable at
localising small objects. We selected the $50$ best-scoring detections for each image, a number empirically large enough to capture most objects present in the scene. We discarded detections larger than $25\%$ of the image size, as they were likely associated with a large scene area, rather than with an object, and could possibly carry information  detrimental to the relative pose estimation task.

\noindent{}\textbf{Camera selection --} We then sampled image sets, as candidates for the nodes of the graphs. In most of our experiments we used 
graphs composed of eight nodes; we compared the performances of PoserNet on large graphs (125 nodes) and small graphs (eight nodes), and in the paper we show how the latter lead to better performance.  Moreover, in the paper we show how training PoserNet on small graphs still allows it to generalise well on large graphs, making the choice of using small graphs not restrictive.

\noindent{}\textbf{ROIs matching --} To establish which candidate nodes are connected, we matched ROIs across images. We passed each image pair through the pre-trained SuperGlue~\cite{sarlin20superglue} model, generating a set of matched keypoints for the pair of images. Of those keypoints, we kept only the ones contained in the bounding boxes (BB) associated with the ROIs. We then defined as matched all BB pairs that shared at least $15$ matched keypoints, with the keypoints spanning at least $30\%$ of the area of the bounding box. These two criteria were defined to ensure that matching bounding boxes enclose the same scene element, and that the shared scene elements constitute a large chunk of the bounding box. We created one edge between two images in a graph if the images were connected via at least five matched BBs.

\noindent{}\textbf{Relative pose estimation --} For each edge, we provide two estimates of the relative transformation
mapping points between the connected nodes. The estimates were generated using the OpenCV~\cite{opencv_library} implementation of the 5-point algorithm, using as inputs either all matched keypoints, or the centres of the matched BBs.
Keypoint-based initialisation and BB-based initialisation result in estimates with different levels of noise: the keypoints - aside from some noise and possible mismatches - are much closer 
to the ideal input of the 5-point algorithm than the BB centres. Even if the two BBs were tightly fitting the same object in the two views, 
the BB centres would not correspond to projections of the same 3D points; and the ROIs are rarely well centred on the objects. 

\noindent{}\textbf{Graph's topology --} In the previous steps we generated all elements necessary to build our graph (nodes and 
edges), but we now have to check how connected the graph is. The presence of isolated nodes or having multiple disconnected subgraphs is not an issue for PoserNet per se, as the former are ignored and the latter are optimised in parallel. For this reason, we accept the large graphs as they are. In the small-graph case, however, given the limited number of nodes available,
we reject all graphs with isolated nodes or isolated subgraphs.

\section{Implementation Details}
In this section, we discuss implementation details for the third-party code referenced in the paper (MultiReg and EIG-SE3) and for PoserNet. We provide general information on the 
hardware and computational time involved in the experiments, and on the changes implemented in the third-party code we used in our
experiments.

\textbf{PoserNet.} 
The MLPs used in PoserNet for updating the embedding of the edges ($\Psi_e$) and of the nodes ($\Psi_n$) consist in three fully connected layers, each followed by leaky ReLUs. The number of units in each layer is $32$. We employed the Adam optimiser, with a learning rate of $10^{-3}$ for the experiments training on small graphs, and of $10^{-2}$ for the experiments training on large graphs. The difference was motivated by the different speed of learning we observed in the two cases. We adopted a learning rate scheduler, which reduced the value of the learning rate after three epochs without improvements on the validation loss. 
Regarding $\alpha$, the parameter used to tune the strength of the different components of the loss, we found that the best results are obtained by setting $\alpha =0.1$.

\textbf{MultiReg.} 
To use MultiReg ~\cite{yew2020-RobustSync} in our experiments, we implemented as few changes as possible to the original code. First,
we added data loaders to process our 7-Scenes graphs; this narrowly follows the process used by the original MultiReg method to process the ScanNet dataset.
Unlike in the original method, we do not augment data by randomly sub-sampling each graph; the ScanNet training data used in the original paper was composed by $60$ graphs, each containing 
from $50$ to over $100$ nodes. Our training set, instead, contains $14k$ graphs of $8$ nodes each, making the augmentation step unnecessary, if not detrimental.
As a last change, we relaxed the conditions for classifying an edge as outlier: the original work labels as outliers and it ignores all graph edges with an associated rotation error of 
over $15\degree$ or a translation error of more than $0.15$ m. We removed this constraint, as it would lead us to ignore a large number of edges.  
Training MultiReg on the eight-nodes sequences took on average 12 hours on a GeForce RTX 2080 Ti with 11 GB of RAM, while evaluation on the test set required less than one minute.

\textbf{EIG-SE3:}
We ran the EIG-SE3 algorithm using the Matlab code originally released by the authors of~\cite{arrigoni2016spectral}, with no modification and just looping over all testing graphs.
This process does not require training, and processing the testing set ($7k$ $8$-nodes graphs) requires approximately $50$ minutes. This is significantly longer than MultiReg's evaluation 
time, though the process could be made more efficient, by parallelising it and evaluating in batches.

\begin{figure}[th!]
   \centering
   \begin{subfigure}[b]{0.49\columnwidth}
     \includegraphics[width=\columnwidth]{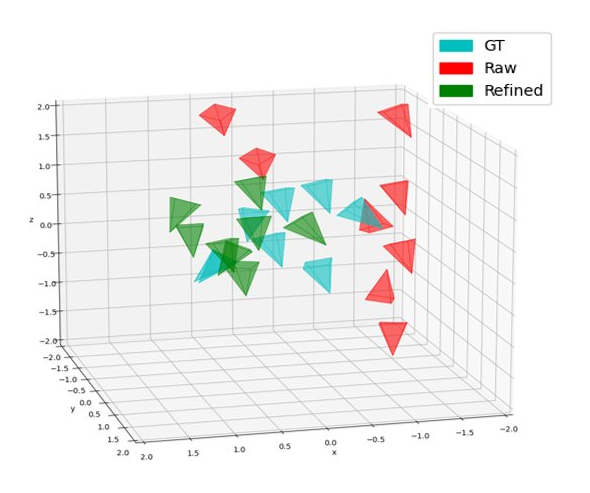}
     \caption{BB-based initialisation}
   \end{subfigure}
   \begin{subfigure}[b]{0.49\columnwidth}
     \includegraphics[width=\columnwidth]{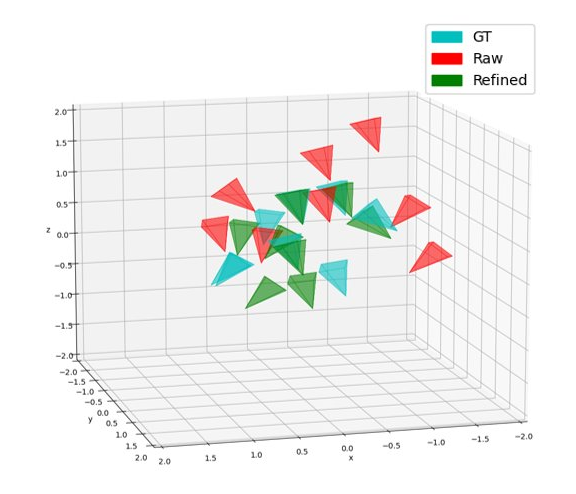}
     \caption{Keypoint-based initialisation}
   \end{subfigure}

   \caption{Effects of PoserNet on the motion averaging output of EIG. We compare ground-truth camera poses (azure) with the output of EIG initialised with the raw relative camera poses (red) and the ones refined with PoserNet (green). For relative poses obtained with both the bounding-box (a) and key-point (b) initialisation method, EIG's output is significantly improved by using PoserNet to refine the input relative poses.}
   \label{fig:eig_output}
\end{figure}

We provide examples of the absolute camera poses generated by EIG in Figure~\ref{fig:eig_output}, comparing
the poses obtained starting from relative poses with and without PoserNet refinement. In the example we can see how PoserNet can 
lead to significantly more accurate poses, both for relative poses initialised using matched keypoints or the matched detection BB centres. We can see how, in the case of BB-based initialisation, using the raw input results in camera poses so noisy that EIG cannot properly align them to the ground truth poses. We also point out that the poses obtained starting from the PoserNet-refined inputs have the correct orientation, and the discrepancies with respect to the ground truth poses are mostly limited to translations. This is not surprising, as the results in Tables~\ref{tab:results_posernet_bb_vs_keypoints},\ref{tab:results_7scenes} of the main paper show how PoserNet improvement of the relative poses is mostly noticeable in the relative rotations rather than in the translation direction.

\section{Highlighting advantages of PoserNet}
\begin{figure}[t]
   \centering
   \begin{minipage}[b]{\columnwidth}
     \includegraphics[width=\columnwidth]{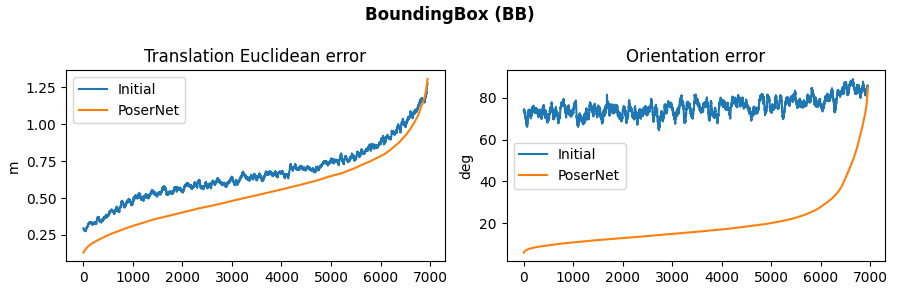}
   \end{minipage}
   \hfill
   \begin{minipage}[b]{\columnwidth}
     \includegraphics[width=\columnwidth]{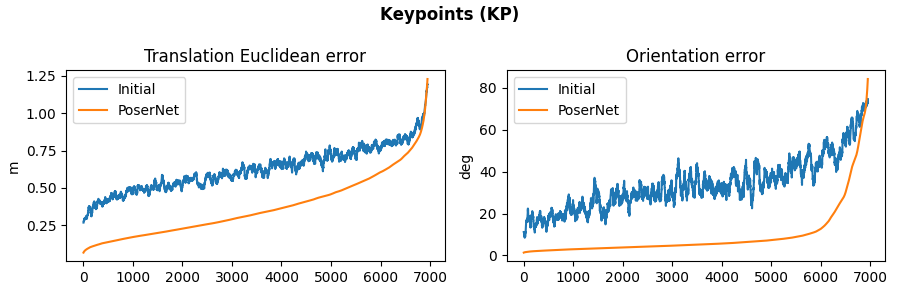}
   \end{minipage}
   \caption{EIG-SE3 Motion Averaging performances on raw inputs and on inputs refined by PoserNet. There is a substantial error reduction both for relative poses computed from the detection centres (Top) and keypoints (Bottom).}
   \label{fig:eig}
\end{figure}

To better highlight the significant benefit of refining the relative pose with PoserNet, in Fig.~\ref{fig:eig} we provide plots of the rotation and translation error distributions over all eight-node testing graphs. To increase interpretability, we sorted the sequences in ascending order of error on the PoserNet-refined predictions. Moreover, given the large number of sequences and likely presence of outliers, we smooth the curves using a rolling average over a window of $50$ graphs. 

\section{Ablation on the depth of PoserNet}
To select the optimal number of rounds of message passing (characterised as GNN ``depth'') for the GNN in PoserNet, we trained four versions of PoserNet with depths ranging from two to five. For this experiment 
we trained each model for the same amount of time (24 hours), initialising the edges of the small graph dataset with the keypoint-based relative poses.
As it can be seen in Table~\ref{tab:gnn_depth}, the highest accuracy is achieved with a GNN of depth $2$, which is the value we used for all the remaining experiments described in the main document.

\begin{table}
\begin{center}
\begin{tabularx}{0.7\textwidth}{X c c c c}
 \toprule
  & 2-level & 3-level & 4-level & 5-level \\ 
  \midrule
 Relative orientation error  &  \textbf{7.5} &  7.6  &  9.4 & 11.2 \\ 
 Translation direction error & \textbf{14.7} & 14.9  & 16.1 & 17.2 \\ 
 \bottomrule
\end{tabularx}
\end{center}
\caption{\label{tab:gnn_depth} Median performance of PoserNet as a function of GNN depth, errors expressed in degrees. Best performance (in bold) was achieved with a two-level GNN.}
\end{table}

\section{Ablation on the keypoint descriptors}

In all our experiments, the initial relative poses are obtained applying the OpenCV 5-point algorithm to either the 
2D centres of matched bounding boxes, or keypoints SuperGlue. These two choices represents opposite use cases,
with the BB centres providing a very rough initialisation (even if the bounding boxes are correctly matched, their centres
are not guaranteed to correspond to the same 3D point), and SuperPoint+SuperGlue being the state-of-the-art for
keypoint detections and matching. We assumed that any other common choice of keypoint matching would result in input relative
pose accuracy subsumed in the range of values covered by our two initialisation strategies, and that showing 
how PoserNet can significantly improved on both could be generalised to the other cases. To support this point,
we show in Table~\ref{keypointscompared} how estimating the relative pose starting from points matched via SIFT or ORB 
is more accurate than using BB centres, and less accurate than using SuperPoint + SuperGlue.
Moreover, we must point out that while using ORB provided better results than SIFT, the ORB matches caused the 5-point 
algorithm to fail in $30\%$ of the image pairs. 
\begin{table}
\begin{center}
\begin{tabular}{c| c c c c }
 \toprule
Metric & BB & SIFT & ORB$^*$ & GLUE \\
\midrule
R & 96.48 & 53.53 & 74.84 & \bf{36.27} \\
t & 89.30 & 90.73 & 90.20 & \bf{87.23} \\
\bottomrule
\end{tabular}
\end{center}
\caption{Median rotation (R) and translation (t) error using SIFT, ORB, SuperPoint+SuperGlue (GLUE), and matched BB centre (BB) to compute the relative poses. $*$ Failed on $30\%$ of pairs.\label{keypointscompared}}
\end{table}

\section{Generalisation of PoserNet to novel scenes}
To assess PoserNet generalisation capabilities, we evaluate on each of the seven scenes a model trained exclusively on the other six. 
For these tests, we consider only eight-node graphs, and we use the key-point based relative pose initialisation approach. The results of this leave-one-out scheme, reported in Table~\ref{tab:loo_test}, suggest PoserNet can generalise well to novel scene: averaging the performances over all seven tests results in average rotation and translation orientation errors compatible with the those obtained training on the full dataset.

\begin{table}
\begin{center}
\begin{tabularx}{\textwidth}{X | c c c c c c c | c | c}
 \toprule
 & Office & Pumpkin & RedKitchen & Stairs & Chess & Fire & Heads & Median & All \\
  \midrule
  Rotation & 5.05 & 9.93 & 7.33 & 9.39 & 7.27 & 8.36 & 8.96 & 8.04 & 7.31 \\
  Translation & 10.59 & 13.05 & 16.82 & 18.07 & 21.03 & 11.15 & 47.54 & 19.75 & 14.54 \\
 \bottomrule
\end{tabularx}
\end{center}
\caption{\label{tab:loo_test} Median performance of PoserNet on each scene when trained only on the other six. The overall performance over the seven tests (Median) is comparable with the performances obtained training PoserNet on the full dataset (All).}
\end{table}

\end{document}